\documentclass{article}
\usepackage{spconf,amsmath,graphicx,hyperref,tabularx}
\usepackage{multirow} 
\usepackage{amssymb}
\usepackage{graphicx}   
\usepackage{xcolor}    
\usepackage{colortbl}   
\usepackage{array}  
\usepackage{multirow}  
\usepackage{diagbox} 
\usepackage{booktabs} 
\usepackage{caption} 
\usepackage{subcaption} 
\usepackage{tabularx}
\definecolor{mygray}{RGB}{230, 230, 230}
\definecolor{bordercolor}{RGB}{215,215,215}
\definecolor{fillcolor}{RGB}{215,215,215}
\usepackage[T1]{fontenc}

\usepackage{graphicx,verbatim}
\usepackage{pifont}
\usepackage{booktabs}

\def\L{{\cal L}}
\title{SSCM: A SPATIAL-SEMANTIC CONSISTENT MODEL FOR MULTI-CONTRAST MRI SUPER-RESOLUTION}
\name{
    Xiaoman Wu\textsuperscript{1}, 
    Lubin Gan\textsuperscript{1},  
    Siying Wu\textsuperscript{2$\star$}, 
    Jing Zhang\textsuperscript{2}\thanks{$\star$ Corresponding author.},
    Yunwei Ou\textsuperscript{3},
    Xiaoyan Sun\textsuperscript{1, 2$\star$}
}

\address{
\textsuperscript{1} University of Science and Technology of China, Anhui, China\\
\textsuperscript{2} Anhui Province Key Laboratory of Biomedical Imaging and Intelligent Processing\\
Institute of Artificial Intelligence, Hefei Comprehensive National Science Center, Anhui, China \\
\textsuperscript{3} Beijing Tiantan Hospital, Capital Medical University, Beijing, China
}
\begin{document}
%
\maketitle

\begin{abstract}
Multi-contrast Magnetic Resonance Imaging super-resolution (MC-MRI SR) aims to enhance low-resolution (LR) contrasts leveraging high-resolution (HR) references, shortening acquisition time and improving imaging efficiency while preserving anatomical details. The main challenge lies in maintaining spatial-semantic consistency, ensuring anatomical structures remain well-aligned and coherent despite structural discrepancies and motion between the target and reference images. Conventional methods insufficiently model spatial–semantic consistency and underuse frequency-domain information, which leads to poor fine-grained alignment and inadequate recovery of high-frequency details. In this paper, we propose the Spatial-Semantic Consistent Model (SSCM), which integrates a Dynamic Spatial Warping Module for inter-contrast spatial alignment, a Semantic-Aware Token Aggregation Block for long-range semantic consistency, and a Spatial-Frequency Fusion Block for fine structure restoration. Experiments on public and private datasets show that SSCM achieves state-of-the-art performance with fewer parameters while ensuring spatially and semantically consistent reconstructions.
\end{abstract}


\section{Introduction}
\label{sec:intro}

\begin{figure}[t]
    \includegraphics[width=\linewidth]{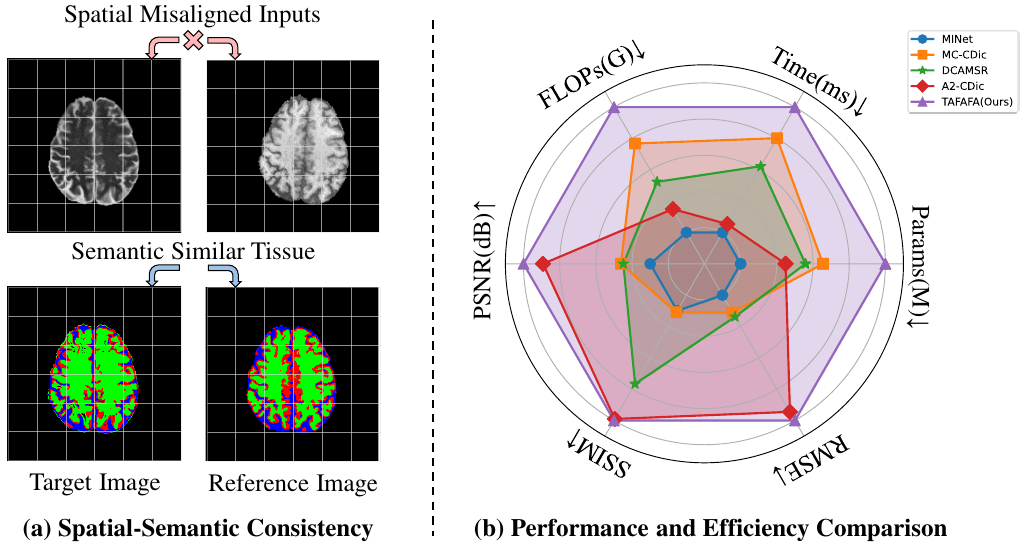}
    \caption{(a) Spatial-semantic consistency. (b) SSCM achieves higher computational efficiency and better performance (lower-is-better metrics are reported after reflection).}
    \label{fig:performance}
\end{figure}

Magnetic Resonance Imaging (MRI) is a widely used non-invasive technique in clinical practice \cite{de2016accuracy, song2019coupled, zheng2025towards,zheng2024odtrack,zheng2025decoupled,zheng2023toward,zheng2022leveraging}, crucial for diagnosing neurological disorders, tumors, and other pathologies by providing complementary tissue information across protocols. Multi-contrast (MC) MRI, such as T1-weighted, T2-weighted, and proton-density (PD) images, is routinely acquired to capture distinct anatomical and pathological features. However, high-resolution (HR) acquisition is limited by long scan times, causing patient discomfort, motion artifacts, and reduced signal-to-noise ratio (SNR) \cite{lyu2020multi, feng2022multimodal,lu2024mace,lu2023tf,lu2024robust,li2025set,gao2024eraseanything,ren2025all,yu2025visual,lu2022copy,yang2025temporal,zhu2024oftsr,1,2,3,4,5,6,7,8}. MC-MRI super-resolution (SR) leverages an HR reference scan to guide the reconstruction of low-resolution (LR) images \cite{li2023ntire,ren2024ninth,wang2025ntire,peng2020cumulative,wang2023decoupling,peng2024lightweight,peng2024towards,wang2023brightness,peng2021ensemble,ren2024ultrapixel,yan2025textual,peng2024efficient,conde2024real,peng2025directing,peng2025pixel,peng2025boosting,he2024latent,di2025qmambabsr,peng2024unveiling,he2024dual,he2024multi,pan2025enhance,wu2025dropout,jiang2024dalpsr,ignatov2025rgb,du2024fc3dnet,jin2024mipi,sun2024beyond,qi2025data,feng2025pmq,xia2024s3mamba,pengboosting,suntext,yakovenko2025aim,xu2025camel,wu2025robustgs}, thereby reducing scan time while improving image quality. In the MC-MRI SR task, maintaining spatial consistency between target and reference images, as well as semantic consistency across different contrasts, is crucial for achieving accurate representation of anatomical structures and key tissues. This enhances the fidelity of HR images, as shown in Fig. \ref{fig:performance} (a).

However, existing methods \cite{wei2025learning,hu2024subdivision,lei2025robust,mao2023disc,li2024rethinking,huang2023accurate,chen2024towards,chen2025mixnet,wu2025dropout,wu2025adaptive,wu2024rethinking,Li2023FCDFusion,Li_2025_CVPR,yi2021structure,yi2021efficient,yi2025fine,wang2025vastsd,wang2024cardiovascular,wang2025angio,gan2025enhancing,gan2025semamil} still face challenges in preserving spatial-semantic consistency. Some methods divide images into local patches and transform them into sequences \cite{lei2025robust, lei2023deep}, which often fragment semantically coherent structures and weaken global contextual understanding. In addition, some methods overlook explicit spatial alignment, making it difficult to guarantee spatial consistency \cite{mao2023disc, li2024rethinking}. Adaptive strategies, such as DCAMSR \cite{huang2023accurate}, use non-local modules to mitigate misalignment but lack fine-grained alignment and rely solely on spatial-domain processing, thereby limiting high-frequency detail recovery and overall reconstruction quality.

\begin{figure*}[t]
    \centering
    \includegraphics[width=\textwidth]{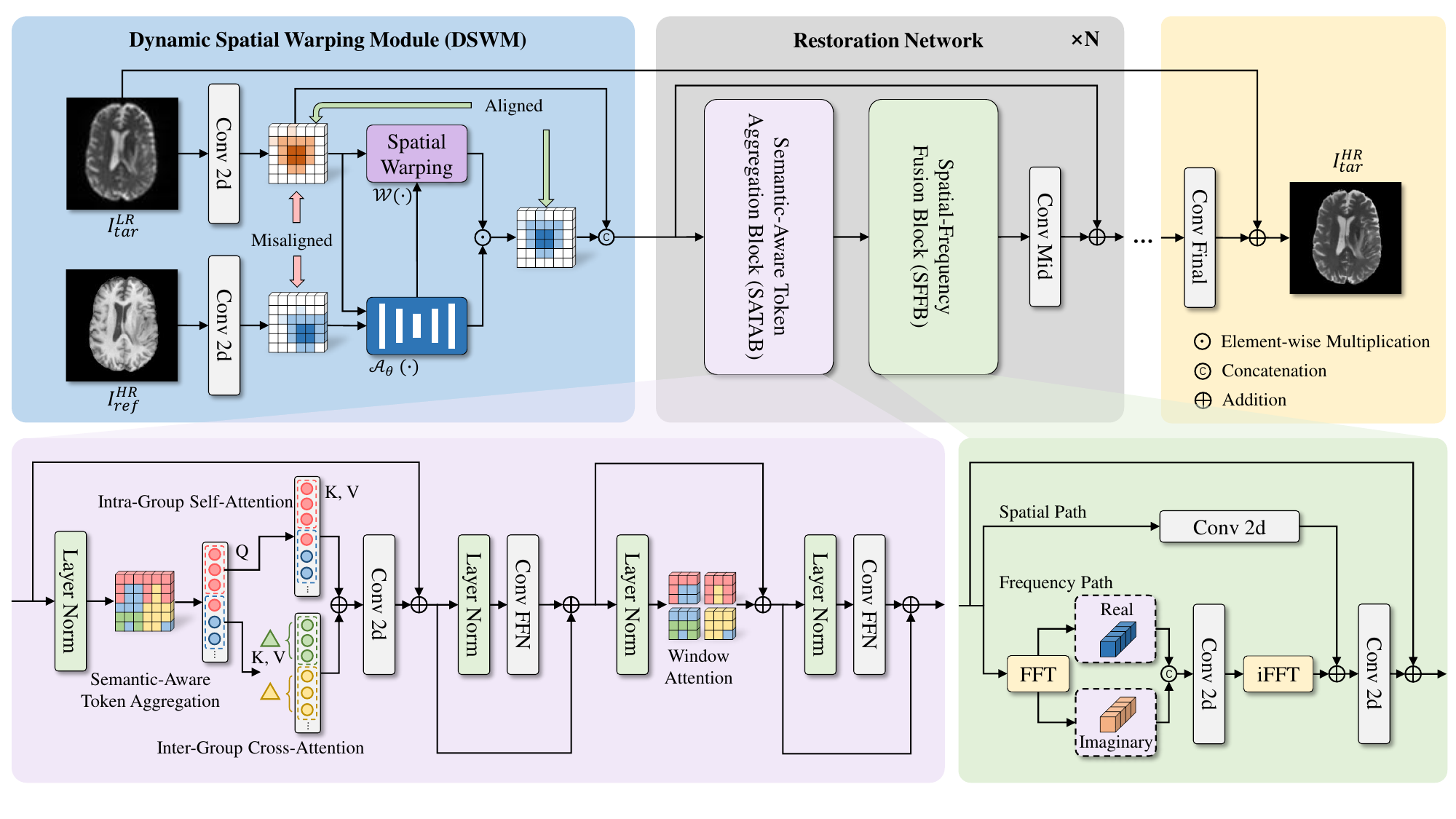}
    \caption{The overall architecture of proposed Spatial-Semantic Consistent Model (SSCM). The model begins with a Dynamic Spatial Warping Module (DSWM), followed by a restoration network of $N$ stacked restoration blocks, each comprising a Semantic-Aware Token Aggregation Block (SATAB) and a Spatial-Frequency Fusion Block (SFFB).}
    \label{fig:mainframe}
\end{figure*}

In this paper, we propose a Spatial-Semantic Consistent Model (SSCM) for MC-MRI SR task. The core of SSCM includes: (1) Dynamic Spatial Warping Module, which corrects inter-scan motion through fine-grained feature alignment and spatial warping; (2) Semantic-Aware Token Aggregation Block, which captures long-range anatomical dependencies and aggregates semantically related tokens, thereby ensuring cross-contrast semantic consistency, preserving structural coherence, and filtering out contrast-inconsistent information; (3) Spacial-Frequency Fusion Module, using a dual-branch design to simultaneously enhance local spatial textures and global spectral information, complementing spatial cues to improve reconstruction quality.

\section{Methods}
\label{sec:majhead}


\subsection{Overall Architecture and Problem Formulation}
\label{ssec:subhead}

Given a low-resolution (LR) input $I_{\text{tar}}^{\text{LR}} \in \mathbb{R}^{H\times W}$ and its corresponding high-resolution (HR) reference $I_{\text{ref}}^{\text{HR}} \in \mathbb{R}^{H\times W}$, our objective is to learn a mapping function that reconstructs a high-quality target $\hat{I} _{\text{tar}}^{\text{HR}} \in \mathbb{R}^{H\times W} $ image .The overall process can be formulated as:

\begin{equation}
    \hat{I}_{\text{tar}}^{\text{HR}} = F(I_{\text{tar}}^{\text{LR}},I_{\text{ref}}^{\text{HR}}).
\end{equation}

As illustrated in Fig. \ref{fig:mainframe}, the proposed SSCM first employs a Dynamic Spatial Warping Module (DSWM). This module extracts features from both the target and reference images while simultaneously correcting for inter-scan motion to achieve fine-grained spatial alignment. Subsequently, a restoration network reconstructs the low-resolution (LR) images to high-resolution (HR) quality. The restoration network consists of $N$ stacked restoration blocks, each composed of a Semantic-Aware Token Aggregation Block (SATAB) and a Spatial-Frequency Fusion Block (SFFB), enabling progressive refinement of degraded features across multiple hierarchies.

\subsection{Dynamic Spatial Warping Module}

We use two convolutional layers to extract features $f_{\text{tar}}$ and $f_{\text{ref}}$, respectively. Then, to correct inter-scan motion, we design a Dynamic Spatial Warping Module (DSWM). Specifically, we introduce a function $\mathcal{A}_{\theta}$ parameterized by $\theta$, which takes $f_{\text{tar}}$ and $f_{\text{ref}}$ as input and predicts a dense 2D displacement field $\bigtriangleup p \in \mathbb{R}^{H\times W\times 2}$. Here, $\bigtriangleup p$ models the pixel-wise spatial offset from the target contrast to the reference.

The reference feature map $f_{\text{ref}}$ is then warped using this displacement field via a differentiable bilinear sampling operation, denoted by $\mathcal{W}$, to produce an aligned feature $f_{\text{ref}}^{\text{aligned}}$:

\begin{equation}
    f_{\text{ref}}^{\text{aligned}}=\mathcal{W}(f_{\text{ref}},\bigtriangleup p=\mathcal{A}_{\theta}(f_{\text{tar}},f_{\text{ref}})).
\end{equation}
The target features $f_{\text{tar}}$ and the aligned reference features $f_{\text{ref}}^{\text{aligned}}$ are concatenated along the channel dimension and then fused using a 1x1 convolution:

\begin{equation}
    f_{\text{in}}=F_{\text{conv}}^{\text{fuse}}(\text{concat}[f_{\text{tar}},f_{\text{ref}}^{\text{aligned}}])\in \mathbb{R}^{C\times H\times W}.
\end{equation}

\label{ssec:subhead}

\subsection{Semantic-Aware Token Aggregation Block}
\label{ssec:subhead}

To efficiently model long-range dependencies and accurately restore brain structure, we employ a Semantic-Aware Token Aggregation Block (SATAB), which performs global attention in a semantic-aware manner. This process avoids the quadratic complexity of standard global self-attention by token aggregation. The input feature map $f_{\text{in}}$ is first reshaped into a sequence of $N=H\times W$ tokens, $\{x_{i}\}_{i=1}^{N}$, where $x_{i}\in \mathbb{R}^{C}$. The core idea is to aggregate these tokens based on their cosine similarity to a set of $K$ learnable token centers, $C=\{c_{k}\}_{k=1}^{K}$, where $c_{k}\in \mathbb{R}^{C}$. These centers are shared across the dataset and updated via exponential moving average (EMA) during training. Each token $x_{i}$ is assigned to a group $G_{k}$ corresponding to the most similar prototype:
\begin{equation}
    \text{Group}(x_{i})=G_{k} ,\space k=\arg\max_j(\frac{x_{i}\cdot c_{j}}{||x_{i}||\cdot ||c_{j}||}).
\end{equation}
Then, we manually split $G_{k}$ into sub-groups $S_k$ of equal size to enhance parallel computing. With tokens now organized into meaningful sub-groups, we perform two types of attention. One of them is intra-group self-attention (SA), with standard multi-head self-attention (MHSA) computed within each sub-group $S_{k}$:
\begin{equation}
    Y_{\text{SA}} = \text{MHSA}(S_kW_A^Q, S_kW_A^k, S_kW_A^V),
\end{equation}
where $W_A^Q,W_A^K$ and $W_A^V$ are weight metrics. Another is inter-group cross-attention (CA). To facilitate global information exchange between groups, the tokens in each sub-group $S_{k}$ (as Queries) attend to the set of global content prototypes $M$ (as Keys and Values):
\begin{equation}
    Y_{\text{CA}} = \text{MHSA}(S_kW_R^Q, CW_R^K, CW_R^V),
\end{equation}
where $W_R^Q,W_R^K$ and $W_R^V$ are weight metrics. The output of attention map, $f_{\text{A}}$, is obtained by:
\begin{equation}
    f_{\text{A}}=F_{\text{conv}}(Y_{\text{SA}},Y_{\text{CA}}).
\end{equation}

Inspired by Liu et al. \cite{liu2023coarse}, we introduce patch-level window attention to refine textural details. The feature map $f_{\text{A}}$ is divided into overlapping patches ${p_j}$ via a sliding window of size $p_s \times p_s$ with stride. Standard MHSA is applied independently to each patch with shared Q, K, V projections:
\begin{equation}
p_{j}^{'}=\text{MHSA}(p_j).
\end{equation}
The processed patches ${p_{j}^{'}}$ are reassembled, averaging overlaps for smoothness. The result is passed through an FFN convolution to produce the final output of SATAB, $f_{\text{SATAB}}$.

\subsection{Spatial-Frequency Fusion Block}
\label{ssec:subhead}

To recover high-frequency details, we propose a Spatial-Frequency Fusion Block (SFFB), which processes input $f_{\text{SATAB}}$ via two parallel paths.
The \textbf{Spatial Path} applies 3×3 convolutions:
\begin{equation}
X_{\text{spat}} = F_{\text{conv}}^{\text{spat}}(f_{\text{SATAB}}).
\end{equation}
The \textbf{Frequency Path} applies the Real Fast Fourier Transform (RFFT), denoted as $\mathcal{F}(\cdot)$, to map features to the frequency domain, modulates them with a $1\times1$ convolution, and then transforms them back using $\mathcal{F}^{-1}$:
\begin{equation}
X_{\text{freq}} = \mathcal{F}^{-1}(F_{\text{conv}}^{\text{freq}}(\text{concat}[\Re, \Im])),
\end{equation}
where $\Re$ and $\Im$ represent the real and imaginary part of $\mathcal{F}(f_{\text{SATAB}})$. The outputs are fused and added to the input:
\begin{equation}
f_{\text{SFFB}} = f_{\text{SATAB}} + F_{\text{conv}}^{\text{fuse}}(X_{\text{spat}} + X_{\text{freq}}).
\end{equation}

For the $n$-th restoration block, combining SATAB ($\mathcal{S}$) and SFFB ($\mathcal{B}$):
\begin{equation}
f_n = f_{n-1} + F_{\text{conv}}^{\text{mid}_n}(\mathcal{B}(\mathcal{S}(f_{n-1}))).
\end{equation}
This residual design refines features progressively while preserving stable information flow.
The HR image is then reconstructed by adding the LR image with the processed feature $f_N$:
\begin{equation}
I_{\text{tar}}^{\text{HR}} = I_{\text{tar}}^{\text{LR}} + F_{\text{conv}}^{\text{final}}(f_N).
\end{equation}

\begin{table*}[!t]
\centering
\caption{Quantitative results on three datasets. Best and second-best results are highlighted with \textbf{bold} and \underline{underline}.}

\scriptsize
\renewcommand{\arraystretch}{1} 
\setlength{\tabcolsep}{2pt}

\begin{tabularx}{\textwidth}{ l | c |  >{\centering\arraybackslash}X >{\centering\arraybackslash}X >{\centering\arraybackslash}X 
| >{\centering\arraybackslash}X >{\centering\arraybackslash}X >{\centering\arraybackslash}X 
| >{\centering\arraybackslash}X >{\centering\arraybackslash}X >{\centering\arraybackslash}X }

\toprule
\multirow{2}{*}{Methods} & \multirow{2}{*}{Params} & \multicolumn{3}{c|}{BraTS 2021 (T1$\rightarrow$T2)} & \multicolumn{3}{c|}{IXI (PD$\rightarrow$T2)} & \multicolumn{3}{c}{External (T1$\rightarrow$T2)} \\
 & &   PSNR$\uparrow$ & SSIM$\uparrow$ & RMSE$\downarrow$ & PSNR$\uparrow$ & SSIM$\uparrow$ & RMSE$\downarrow$ & PSNR$\uparrow$ & SSIM$\uparrow$ & RMSE$\downarrow$ \\
\midrule
ZP(zero-padding)                   & --    & 31.14$\pm$2.56 & 0.8129$\pm$0.03 & 6.77$\pm$1.69 & 32.07$\pm$2.08 & 0.8722$\pm$0.02 & 5.98$\pm$1.69 & 35.31$\pm$3.14 & 0.8971$\pm$0.02 & 4.77$\pm$1.34 \\
SwinIR \cite{liang2021swinir}   & 11.9M & 36.57$\pm$1.44 & 0.9782$\pm$0.01 & 3.92$\pm$1.21 & 35.99$\pm$5.15 & 0.9562$\pm$0.02 & 4.38$\pm$1.46 & 36.24$\pm$4.18 & 0.9667$\pm$0.02 & 4.15$\pm$1.22 \\
MINet \cite{feng2021multi}      & 11.9M & 38.17$\pm$1.29 & 0.9700$\pm$0.01 & 3.25$\pm$1.08 & 40.69$\pm$4.38 & 0.9691$\pm$0.01 & 2.16$\pm$0.72 & 36.59$\pm$4.06 & 0.9706$\pm$0.02 & 3.99$\pm$1.09 \\
MC-CDic \cite{lei2023deep}      & 8.6M  & 39.29$\pm$2.01 & 0.9801$\pm$0.01 & 3.17$\pm$0.74 & 41.15$\pm$2.87 & 0.9821$\pm$0.03 & 2.50$\pm$0.84 & 36.86$\pm$2.48 & 0.9751$\pm$0.01 & 3.85$\pm$1.03 \\
DCAMSR \cite{huang2023accurate} & 9.3M  & 39.28$\pm$1.47 & 0.9848$\pm$0.01 & 3.15$\pm$0.69 & \underline{43.17$\pm$3.83} & 0.9852$\pm$0.01 & \underline{1.99$\pm$0.74} & 36.78$\pm$2.03 & 0.9740$\pm$0.01 & 3.98$\pm$1.25 \\
WavTrans \cite{li2022wavtrans}  & 2.1M  & 39.18$\pm$1.55 & 0.9851$\pm$0.01 & 3.25$\pm$0.71 & 43.12$\pm$4.32 & \underline{0.9859$\pm$0.01} & 2.02$\pm$0.72 & \underline{37.83$\pm$3.74} & 0.9759$\pm$0.01 & \underline{3.72$\pm$0.97} \\
A2-CDic \cite{lei2025robust}    & 10.1M & \underline{39.61$\pm$1.59} & \underline{0.9871$\pm$0.01} & \underline{2.71$\pm$0.85} & 41.21$\pm$2.80 & 0.9783$\pm$0.03 & 2.46$\pm$0.81 & 36.32$\pm$2.66 & \underline{0.9775$\pm$0.01} & 4.08$\pm$1.10 \\
\textbf{SSCM (Ours)}         & 6.1M  & \textbf{39.69$\pm$1.88} & \textbf{0.9872$\pm$0.01} & \textbf{2.67$\pm$0.76} & \textbf{43.40$\pm$4.39} & \textbf{0.9864$\pm$0.01} & \textbf{1.91$\pm$0.77} & \textbf{38.41$\pm$3.58} & \textbf{0.9779$\pm$0.01} & \textbf{3.28$\pm$1.04} \\
\bottomrule
\end{tabularx}
\label{tab:main}
\end{table*}

\begin{figure*}[!t]
    \includegraphics[width=\linewidth]{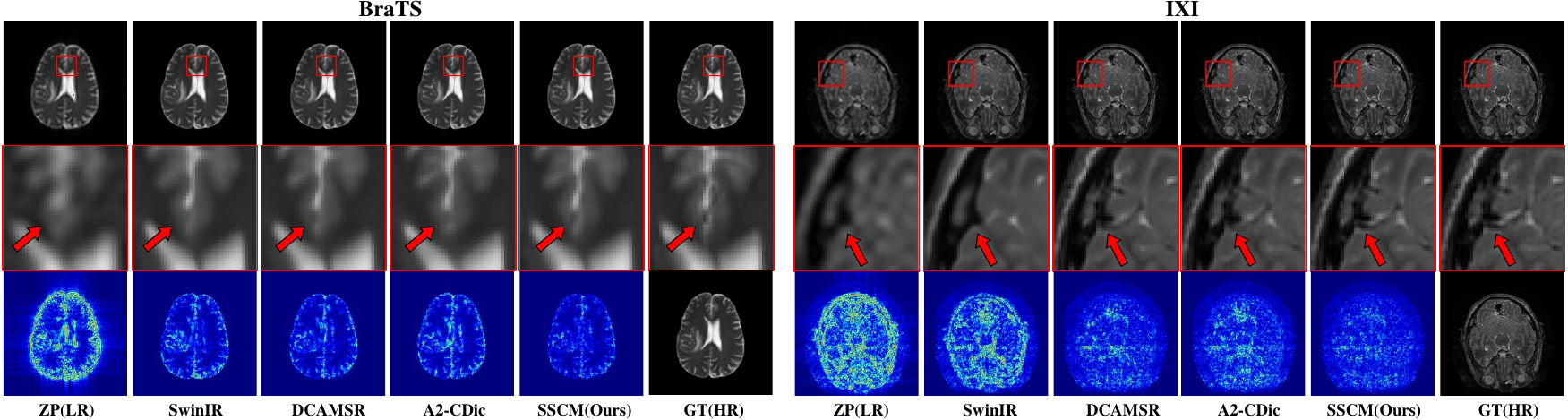}
    \caption{Visual comparisons of different SR methods on the BraTS 2021 and IXI datasets. Zoomed-in regions and heatmaps highlight differences in structural details and reconstruction quality.}
    \label{fig:zoom}
\end{figure*}

\newcommand{\cmark}{\textcolor{green!70!black}{\ding{51}}}
\newcommand{\xmark}{\textcolor{red}{\ding{55}}}

\begin{table}[t]
    \centering
    \caption{Ablation study for DSWM SATAB and SFFB.}
    \renewcommand\arraystretch{1.05}
    \scriptsize
    \setlength{\tabcolsep}{6pt}
    \begin{tabular}{@{} c c c | c c | c c @{}}
        \toprule
        \multicolumn{3}{@{}c|}{\textbf{Our Proposed}}
          & \multicolumn{2}{c|}{BraTS 2021}
          & \multicolumn{2}{c}{IXI} \\
        DSWM & SATAB & SFFB & PSNR$\uparrow$ & SSIM$\uparrow$ & PSNR$\uparrow$ & SSIM$\uparrow$ \\
        \midrule
        \xmark & \xmark & \xmark & 38.22 & 0.9737 & 40.97 & 0.9769 \\
        \xmark & \cmark & \cmark & 39.33 & 0.9841 & 42.76 & 0.9845 \\
        \cmark & \xmark & \cmark & 38.94 & 0.9809 & 42.14 & 0.9821 \\
        \cmark & \cmark & \xmark & 39.25 & 0.9828 & 42.58 & 0.9829 \\
        \cmark & \cmark & \cmark & 39.69 & 0.9872 & 43.40 & 0.9864 \\
        \bottomrule
    \end{tabular}
    \label{tab:abla1}
\end{table}

\section{experiments}
\label{sec:typestyle}

\subsection{Datasets and Implementation Details}
\label{ssec:subhead}

\textbf{Datasets.} We validated our proposed SSCM on three datasets: two public benchmarks BraTS 2021\cite{baid2021rsna}, IXI Dataset \footnote{IXI Dataset, available at \url{https://brain-development.org/ixi-dataset/}} and one private clinical dataset (as the external testing set).
For two public datasets, we partitioned the subjects into training (80\%), validation (10\%), and testing (10\%) sets. The model’s generalization performance was validated on the private clinical dataset using parameters trained on BraTS 2021.

\textbf{Implementation Details.} Our framework was optimized end-to-end using the L1 loss function over $5\times 10^{5} $ iterations. The LR images were generated by k-space center cropping and zero-padding (ZP) \cite{lyu2020multi}, \cite{zhao2019channel}. We evaluated different models on the scaling factor of $\times$4 for SR. For a fair comparison, all state-of-the-art methods were re-trained on all datasets with an initial learning rate of $2\times 10^{-4}$, identical to ours. We employ Peak Signal-to-Noise Ratio (PSNR), Structural Similarity Index (SSIM) and Root Mean Squared Error (RMSE) to evaluate the model's capability. Higher PSNR and SSIM values and lower RMSE values indicate better performance.

\subsection{Results and Comparison}
\label{ssec:subhead}

We evaluated SSCM against one classic single image SR method SwinIR \cite{liang2021swinir}, and a suite of state-of-the-art MC-MRI SR methods, including MINet \cite{feng2021multi}, MC-CDic \cite{lei2023deep}, DCAMSR \cite{huang2023accurate}, WavTrans \cite{li2022wavtrans}, and A2-CDic \cite{lei2025robust}. Fig. \ref{fig:performance} (b) shows that our method provides an efficient and effective solution to the MC-MRI task. 
\textbf{Quantitative Comparison:} Table \ref{tab:main} presents the quantitative results on all three testing sets. Our proposed method consistently achieves the highest PSNR and SSIM scores, and lowest RMSE across all benchmarks, establishing a new state-of-the-art with only 6.1M parameters. Notably, the performance gains are particularly pronounced on the external testing set, demonstrating our model's good generalization ability.
\textbf{Qualitative Comparison:} As shown in Fig. \ref{fig:zoom}, SSCM generates images with superior perceptual quality. The zoomed-in regions and heatmaps (scaled by 10 for clarity) clearly demonstrate our model's ability to restore sharp anatomical edges and fine cortical textures that are lost or over-smoothed by other methods.

\subsection{Ablation Studies}
\label{ssec:subhead}

As shown in Table \ref{tab:abla1}, we conducted ablation studies on two public datasets to validate the efficacy of SSCM. Starting from a baseline with window attention, we incrementally introduced our proposed modules. The incremental addition of DSWM, SATAB, and SFFB yields performance gains of 0.36 dB, 0.75 dB, and 0.44 dB in PSNR on the BraTS dataset, respectively. These results demonstrate that our proposed modules enhance spatial-semantic consistency, leading to improved reconstruction performance.



\section{Conclusion}

In conclusion, we proposed the Spatial-Semantic Consistent Model for MC-MRI SR to address the challenge of preserving spatial–semantic consistency. Our approach integrates a Dynamic Spatial Warping Module, a Semantic-Aware Token Aggregation Block , and a Spatial-Frequency Fusion Block to achieve high-quality image reconstruction. Experimental results on three diverse datasets clearly demonstrate the superiority of our method over existing techniques. Ablation studies further validate the essential contribution of each proposed component. Moreover, our method exhibits strong generalization and parameter efficiency, underscoring its potential for clinical application.

\bibliographystyle{IEEEbib}

\end{document}